\documentclass{article}

\usepackage{arxiv}

\usepackage[utf8]{inputenc} % allow utf-8 input
\usepackage[T1]{fontenc}    % use 8-bit T1 fonts
\usepackage{amsmath,amssymb,amsfonts}
\usepackage{graphicx}
\usepackage{hyperref}
\usepackage{algorithm}
\usepackage{booktabs}   % For professional table formatting (toprule, midrule, bottomrule)
\usepackage{multirow}   % For merged cells across multiple rows
\usepackage{array}      % Enhanced array and tabular environments
\usepackage{amsmath}    % For mathematical formatting
\usepackage{amssymb}    % For additional mathematical symbols
\usepackage{natbib}
\usepackage{graphicx}
\usepackage{algpseudocode}
\usepackage{geometry}
\usepackage{adjustbox}
\usepackage{wrapfig}
\usepackage{authblk}

\title{Curvature-Aware Safety 
Restoration In LLMs Fine-Tuning}

\date{}
\author{
Thong Bach\textsuperscript{1}\thanks{\texttt{t.bach@deakin.edu.au}} \quad
Thanh Nguyen-Tang\textsuperscript{2}
 \quad
Dung Nguyen\textsuperscript{1} \quad
Thao Minh Le\textsuperscript{3} \quad
Truyen Tran\textsuperscript{1}
}

\affil{
\textsuperscript{1}Applied Artificial Intelligence Intiative (A2I2), Deakin University\\
\textsuperscript{2}Department of Data Science, New Jersey Institute of Technology\\
\textsuperscript{3}Pennsylvania State University\\
}

\renewcommand{\shorttitle}{Curvature-Aware Safety 
Restoration In LLMs Fine-Tuning}

\hypersetup{
pdftitle={A template for the arxiv style},
pdfsubject={q-bio.NC, q-bio.QM},
pdfauthor={David S.~Hippocampus, Elias D.~Striatum},
pdfkeywords={First keyword, Second keyword, More},
}

\begin{document}
\maketitle
\begin{abstract}
Fine-tuning Large Language Models (LLMs) for downstream tasks often compromises safety alignment, even when using parameter-efficient methods like LoRA. In this work, we uncover a notable property: fine-tuned models preserve the geometric structure of their loss landscapes concerning harmful content, regardless of the fine-tuning method employed. This suggests that safety behaviors are not erased but shifted to less influential regions of the parameter space. Building on this insight, we propose a curvature-aware alignment restoration method that leverages influence functions and second-order optimization to selectively increase loss on harmful inputs while preserving task performance. By navigating the shared geometry between base and fine-tuned models, our method discourages unsafe outputs while preserving task-relevant performance, avoiding full reversion and enabling precise, low-impact updates. Extensive evaluations across multiple model families and adversarial settings show that our approach efficiently reduces harmful responses while maintaining or even improving utility and few-shot learning performance.
% These results demonstrate that principled navigation of the loss landscape offers a practical path to restoring safety while balancing utility.
\end{abstract}
\section{Introduction}

Large Language Models (LLMs) encode safety-aligned behaviors during pretraining, but these safeguards deteriorate during task-specific fine-tuning, a phenomenon we identify as \textit{safety alignment drift}. Studies demonstrate that even minimal fine-tuning can compromise safety mechanisms, with models like GPT-3.5 Turbo becoming consistently unsafe after adaptation on just 10 adversarial examples~\citep{qi2023finetuning}. Attempts to address this issue by modifying model behavior generally fall into two main categories, both of which suffer from inherent limitations. \textbf{Behavioral unlearning} methods attempt to remove undesirable knowledge or responses~\citep{cao2015, bourtoule2021}, but often require costly retraining or risk catastrophic forgetting. \textbf{Model editing} approaches aim to update factual associations or local behaviors through direct parameter intervention~\citep{meng2022rome, mitchell2022fast}, yet struggle to generalize beyond narrow scopes or isolated prompts. To solve these issues, we propose a new direction that treats safety behavior as an intrinsic property of the model’s geometry and seeks to restore alignment through curvature-aware navigation of the loss landscape.

Our key insight, supported by extensive empirical analysis (Section~\ref{empirical}), is that models preserve notable structural properties in their loss landscapes with respect to harmful content after fine-tuning. Specifically, we observe high correlations in models' responses to harmful inputs before and after fine-tuning, despite substantial divergence in other functional behaviors. This suggests that safety mechanisms remain largely preserved in the parameter space, merely shifted to less dominant regions during task-specific optimization.

This observation motivates our novel approach: \textit{curvature-aware alignment restoration}. We leverage the preserved geometry of the loss landscape to restore safety boundaries. By employing influence functions and second-order optimization techniques, our method navigates the parameter space to increase loss on harmful inputs while minimizing impact on task performance. Our contributions include:
\begin{itemize}
    \item We identify and empirically validate a key insight: Fine-tuning preserves the geometric structure of the loss landscape for harmful content across diverse model families.
    \item We propose a curvature-aware alignment restoration method that leverages influence functions and second-order optimization to suppress harmful behaviors.
    \item We demonstrate that our approach significantly reduces harmful responses while preserving task performance, enhancing few-shot generalization, and improving robustness to adversarial attacks and parameter perturbations.
\end{itemize}

% Our work addresses a fundamental challenge in LLM deployment, enabling practitioners to fine-tune models for specialized domains without compromising critical safety properties.

\section{Empirical Evidence and Loss Landscape Analysis}
\label{empirical}
\begin{table}[t]
\centering
\caption{Pearson correlation coefficients between base and fine-tuned models' responses on harmful content (HEx-PHI), task-specific data (Dolly), and general data (Alpaca). High correlations ($>$0.77) for harmful content across all models indicate preserved safety structure, while low correlations on task/general data show significant behavioral changes during fine-tuning. This asymmetric preservation validates our hypothesis that safety mechanisms remain structurally intact in the loss landscape.}
\label{tab:correlations}
\renewcommand{\arraystretch}{1.2}
\setlength{\tabcolsep}{12pt}
\begin{tabular}{@{}lccc@{}}
\toprule
\textbf{Models} & \textbf{Harmful} & \textbf{Dolly} & \textbf{Alpaca} \\
\midrule
LLama-2 7B     & \textbf{0.992} & 0.056 & -0.055 \\
LLama-2 13B    & \textbf{0.994} & 0.084 & 0.085 \\
LLama 3 8B     & \textbf{0.995} & 0.550 & 0.510 \\
Mistra v3 7B   & \textbf{0.771} & 0.167 & 0.087 \\
Gemma 2        & \textbf{0.799} & 0.291 & 0.199 \\
Qwen 2.5 7B    & \textbf{0.994} & 0.014 & 0.067 \\
\bottomrule
\end{tabular}
\end{table}

In this section, we first present empirical evidence demonstrating high correlations between base and fine-tuned models' responses to harmful content, despite divergence in task performance. We then visualize and quantify this preserved geometry through loss landscape analysis, providing the foundation for our curvature-aware restoration approach.

\subsection{Empirical Validation}
\label{empirical:2.1}
We analyze multiple model families, measuring Pearson correlation coefficients between base and tuned models's response across three distinct data categories: harmful content (HEx-PHI ~\citep{qi2023finetuning}: a benchmark dataset of 330 harmful instructions across 11 policy-based categories), task-specific data (Dolly testset ~\citep{dolly2023}, 200 examples), and general data (Alpaca testset ~\citep{alpaca2023}, 200 examples). 

These correlations quantify how consistently models respond to the same inputs before and after fine-tuning. For each dataset $\mathcal{D}$, we compute the Pearson correlation coefficient:
\begin{equation*}
r = \frac{\sum_{x \in \mathcal{D}}(L_{\text{base}}(x) - \overline{L}_{\text{base}})(L_{\text{tuned}}(x) - \overline{L}_{\text{tuned}})}{\sqrt{\sum_{x \in \mathcal{D}}(L_{\text{base}}(x) - \overline{L}_{\text{base}})^2}\sqrt{\sum_{x \in \mathcal{D}}(L_{\text{tuned}}(x) - \overline{L}_{\text{tuned}})^2}}
\end{equation*}
where $L_{\text{base}}(x)$ and $L_{\text{tuned}}(x)$ are the cross-entropy losses of the base and fine-tuned models on example $x$, and $\overline{L}_{\text{base}}$ and $\overline{L}_{\text{tuned}}$ are their respective mean values across dataset $\mathcal{D}$. Higher correlation indicates the fine-tuned model maintains similar response behavior to the base model, despite parameter changes. By comparing correlations across different input categories, we can detect whether safety-relevant behaviors remain intact despite changes to task-specific capabilities.

\begin{wrapfigure}{r}{0.5\textwidth}
\centering
\includegraphics[width=0.5\textwidth]{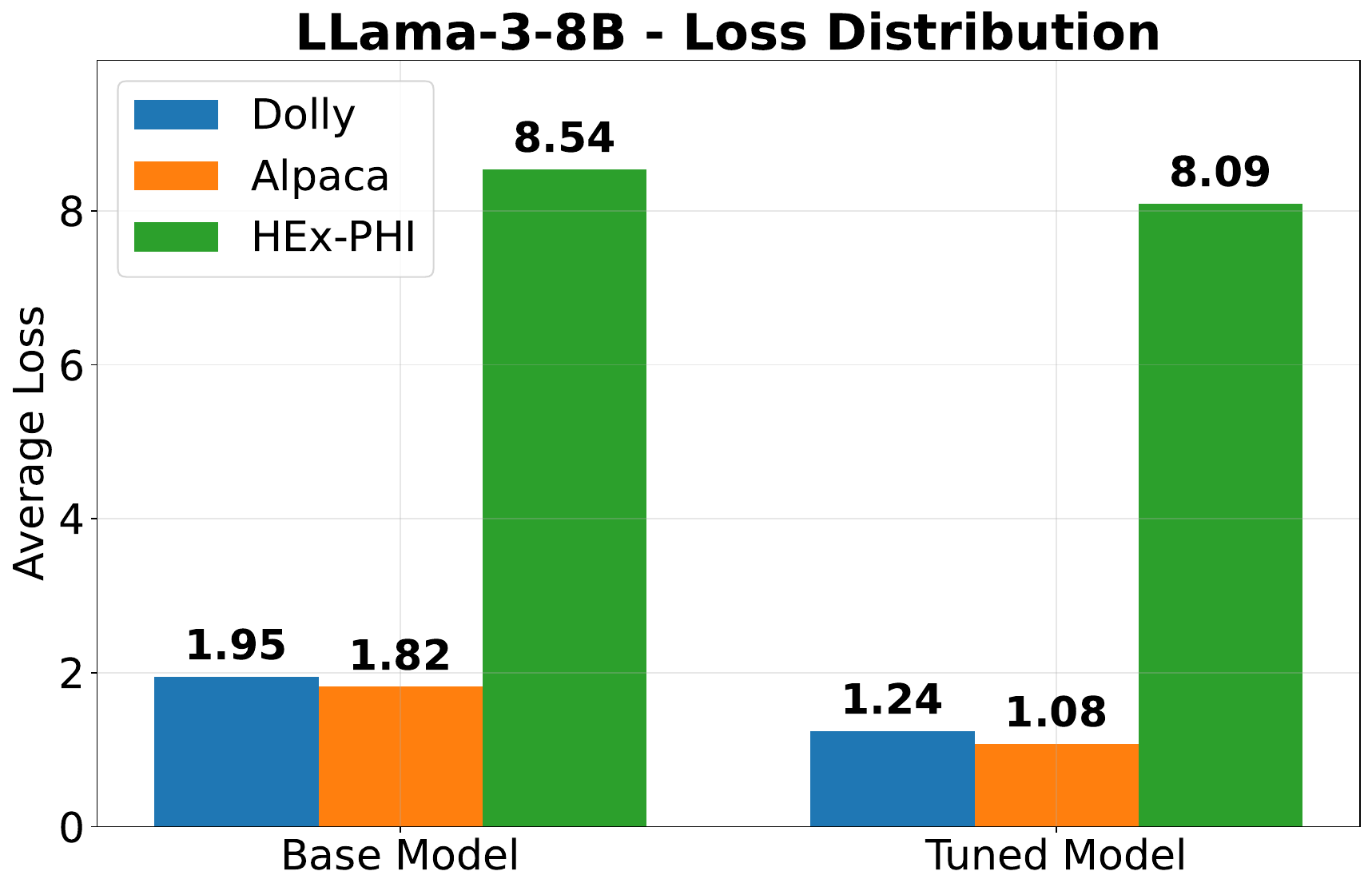}
\caption{Average loss comparison across base and fine-tuned LLama-3 8B Instruct models for three datasets: Dolly (task-specific), Alpaca (general), and HEx-PHI (harmful). Harmful content consistently exhibits higher loss compared to benign content in both model states, showing that harmful content consistently lies in a distinct and preserved region of the loss landscape.}
\label{fig:loss_comparison_single}
\end{wrapfigure}

Generally, our analysis reveals two insights:

\begin{enumerate}
   \item \textbf{Safety response preservation:} In Table \ref{tab:correlations}, we show that despite parameter changes during fine-tuning, models consistently maintain strong response similarity (r > 0.77), contrasting with low or even negative correlations on task and general data. This suggests that safety mechanisms remain structurally unchanged during task-specific optimization.
   
   \item \textbf{Distinct safety regions in loss landscape:} In Figure~\ref{fig:loss_comparison_single} we measure the loss of LLama-3 8B Instruct on these data. Generally, harmful content consistently generates higher loss values ($8.54$ and $8.09$) compared to benign content ($1.82-1.95$ and $1.08-1.24$) in both model states, suggesting potential separation between harmful and task-relevant regions in the loss landscape. More detailed loss analysis will be presented in Appendix \ref{appendix:unlearning}.
\end{enumerate}

Based on these findings, we state our hypothesis: \textbf{safety behaviors exist in a functionally distinct region of the loss landscape that remains largely undisturbed by task-specific fine-tuning}. Therefore, developing the targeted restoration module to recover safety behaviors without compromising useful task capabilities is feasible.
\subsection{Loss Landscape visualization}
\label{section_2:2.2}
To further support our hypothesis, we visualize the loss landscapes of both the base and fine-tuned models using a 3D projection technique. Rather than sampling arbitrary directions in parameter space, we construct perturbation directions informed by gradients computed on harmful and benign inputs. Specifically, we focus on attention and MLP layers, which most strongly influence model behavior. For each model, we generate two approximately orthogonal perturbation vectors ($\mathbf{d}_1$ and $\mathbf{d}_2$) and evaluate the model's loss across a grid ($20 \times 20$) of perturbation magnitudes. We create this grid by varying coefficients $\lambda_1$ and $\lambda_2$ within the range $[-0.01, 0.01]$ and applying the perturbation $\theta_{perturbed} = \theta_{original} + \lambda_1\mathbf{d}_1 + \lambda_2\mathbf{d}_2$ to the model parameters. At each grid point, we compute the loss using a consistent set of 32 validated samples, resulting in a 3D surface where the $x$-axis and $y$-axis represent perturbation magnitudes along each direction, and the $z$-axis shows the corresponding loss value. Full implementation details are provided in Appendix \ref{appendix:loss_visualize}.
\begin{figure}[h]
\centering
\includegraphics[width=\textwidth]{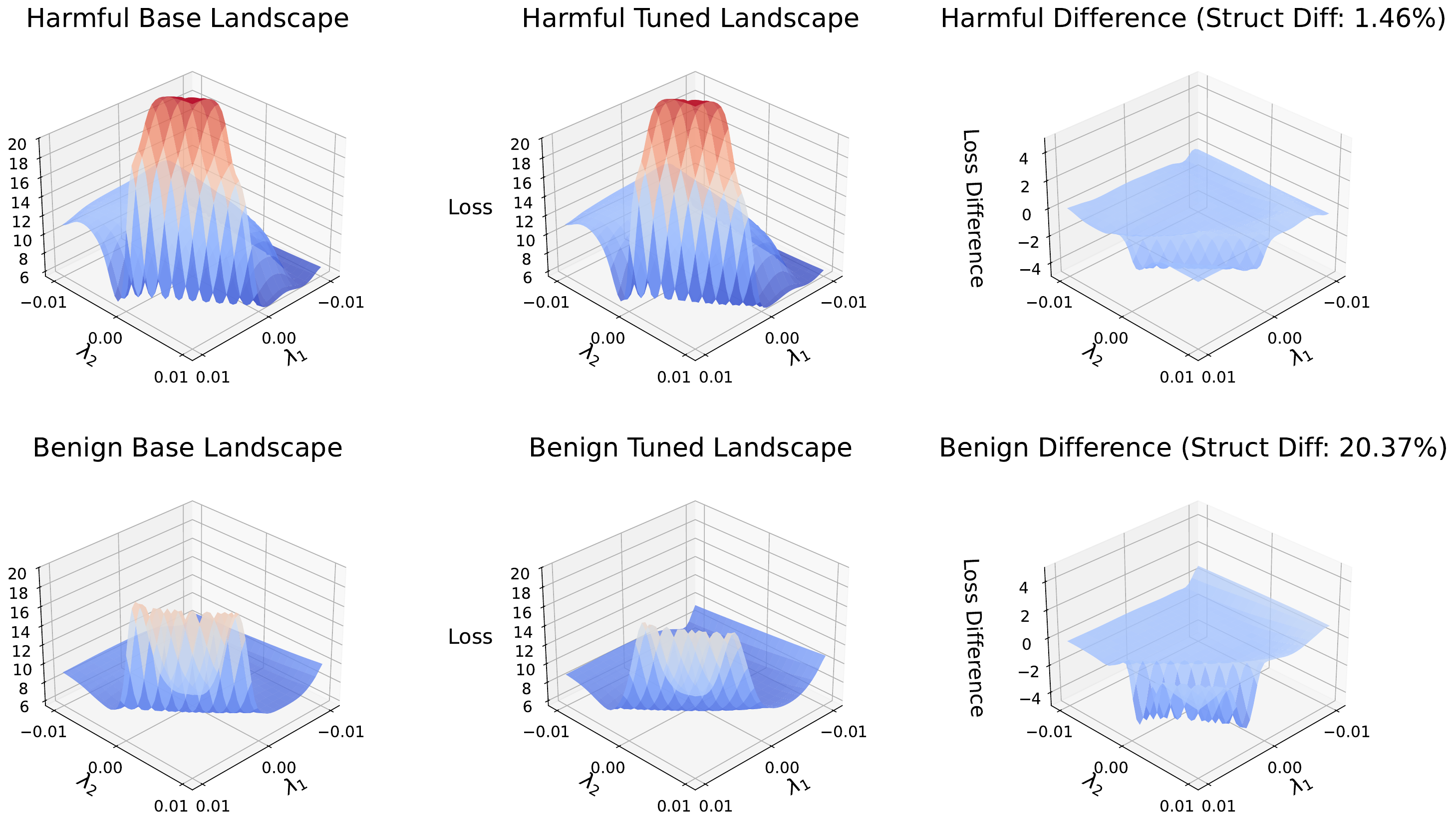}
\caption{3D loss landscape visualization for LLama-3 8B Instruct using gradient-informed direction projection (Section \ref{section_2:2.2}). The top row shows the loss landscape of harmful content (HEx-PHI), while the bottom row shows for general data (Alpaca). Comparison between base (left) and fine-tuned (middle) models reveals preserved topological features for harmful content (structural difference: 1.46\%), while general data landscapes undergo substantial transformation (structural difference: 20.37\%). These quantitative measures of landscape change confirm that safety-relevant regions remain largely undisturbed during task-specific fine-tuning, providing direct evidence for our hypothesis of preserved safety mechanisms.}
\label{fig:loss_landscape}
\end{figure}

The 3D plot in Figure~\ref{fig:loss_landscape} reveals clear evidence for our hypothesis. The loss landscape for harmful content maintains remarkably similar topological features between base and fine-tuned models, with consistent valleys, peaks, and curvature characteristics. We quantify this structural preservation using a correlation-based metric: $\text{StructDiff} = (1 - |\text{corr}(\nabla^2 L_{\text{base}}, \nabla^2 L_{\text{tuned}})|) \times 100\%$, where $\nabla^2 L$ is the Laplacian of the loss landscape. This metric captures differences in curvature patterns rather than absolute loss values, revealing only 1.46\% structural difference for harmful content despite fine-tuning.

In contrast, the loss landscape for general-purpose data changes significantly, exhibiting 20.37\% structural difference in both global geometry and local minima positions.
% While fine-tuning has shifted the harmful content landscape to a slightly lower loss region, its fundamental structure remains largely preserved.
This visualization provides direct confirmation that fine-tuning primarily affects task-specific regions of the parameter space while leaving safety-relevant regions structurally maintained, creating a natural opportunity for targeted safety restoration. Additional results on loss landscapes for LoRA fine-tuning are available in the Appendix.

These visualization results explain the high correlation coefficients documented in Section \ref{empirical:2.1} and establish a foundation for our alignment restoration approach. The idea is to identify and leverage preserved landscape features and use these to navigate toward parameter configurations that maintain task performance while reinforcing safety boundaries.

\section{Curvature-aware alignment restoration}

In this section, we introduce our curvature-aware alignment restoration approach, which provides a principled way to steer a fine-tuned LLM back toward the safety behavior encoded in its base model while preserving desirable task-specific knowledge. Our method is motivated by the shared loss landscape structure between base and tuned models, as demonstrated in our empirical analysis.
\subsection{Problem Formulation and Optimization Approach}

Let us define $\theta_{\text{base}}$ as the parameters of the pretrained, safe base model, and $\theta_{\text{tuned}}$ as the parameters of the fine-tuned model. We use two distinct datasets: 

a \textit{retain set} containing benign, task-relevant examples where performance should be preserved, and a \textit{forget set} containing potentially harmful examples where safety alignment should be restored.

For both datasets, we employ the standard autoregressive language modeling loss:
\begin{equation}
L(x; \theta) = -\sum_{i=1}^{|x|} \log p_{\theta}(x_i | x_{<i})
\end{equation}
where $x$ represents an input sequence and $p_{\theta}(x_i | x_{<i})$ is the model's predicted probability for token $x_i$ given preceding tokens.

Our goal is to update $\theta_{\text{tuned}}$ toward a point $\theta_{\text{updated}}$ that preserves $L_{\text{retain}}$ (the loss on retain set) while increasing $L_{\text{forget}}$ (the loss on forget set). We formulate this as a constrained optimization problem:

\begin{equation}
\label{eq:init_constraint}
\max_{\theta} L_{\text{forget}}(\theta) \quad \text{s.t.} \quad L_{\text{retain}}(\theta) \leq L_{\text{retain}}(\theta_{\text{tuned}}) + \epsilon
\end{equation}

where $\epsilon$ is a small positive scalar allowing limited degradation in retain set performance. Based on extensive empirical validation, we established $\epsilon = 0.1$ as a default constraint threshold, ensuring the recovered model maintains task performance within an acceptable margin of the fine-tuned baseline.

This formulation can be theoretically justified through a second-order Taylor approximation of the retain loss around $\theta_{\text{tuned}}$:

\begin{equation}
L_{\text{retain}}(\theta_{\text{tuned}} + \Delta\theta) \approx L_{\text{retain}}(\theta_{\text{tuned}}) + \nabla L_{\text{retain}}^{\top}\Delta\theta + \frac{1}{2}\Delta\theta^{\top} H_{\text{retain}}\Delta\theta
\end{equation}

Under this approximation, the influence function update provides the steepest descent direction for $L_{\text{forget}}$ in the Riemannian geometry defined by $H_{\text{retain}}$ \citep{amari1998natural}:

\begin{equation}
\label{eq:constraint_max}
\Delta\theta_{\text{influence}} = \arg\max_{\Delta\theta} \nabla L_{\text{forget}}^{\top}\Delta\theta \quad \text{s.t.} \quad \|\Delta\theta\|_{H_{\text{retain}}} \leq \delta
\end{equation}

Here, $\delta > 0$ defines the allowable trust region radius with respect to the local geometry of the retain loss, measured via the Mahalanobis norm $\|\Delta \theta\|_{H_{\text{retain}}} = \sqrt{\Delta \theta^\top H_{\text{retain}} \Delta \theta}$. This parameter is directly related to the constraint threshold $\epsilon$ in Equation \ref{eq:init_constraint}: smaller values of $\delta$ ensure updates remain in regions where the quadratic approximation is valid, thereby helping satisfy the $\epsilon$-bounded retain loss constraint. Intuitively, this constraint ensures that the update direction increases the forget loss without significantly increasing the retain loss, as measured by its local curvature. Solving this constrained optimization yields the steepest ascent direction for $L_{\text{forget}}$ under a Riemannian metric induced by $H_{\text{retain}}$.

Directly solving Equation \ref{eq:constraint_max} may be computationally expensive. Therefore, we adopt a tractable approximation based on influence functions, as shown below:

\begin{equation}
\label{eq:influence_max}
\Delta\theta_{\text{influence}} = H_{\text{retain}}^{-1}\nabla L_{\text{forget}}(\theta_{\text{tuned}})
\end{equation}

% This approximation can be interpreted as the unconstrained solution to Equation \ref{eq:constraint_max}, where the trust region constraint is relaxed. Specifically, Equation \ref{eq:influence_max} represents the steepest ascent direction for $L_{\text{forget}}$ under the curvature geometry of the retain set, without explicitly enforcing a norm constraint. We later enforce trust-region stability through step scaling and L-BFGS curvature filtering (see Appendix \ref{appendix:curvature}).

This approximation can be interpreted as the unconstrained solution to Equation \ref{eq:constraint_max}, where the trust region constraint is relaxed. Specifically, Equation \ref{eq:influence_max} represents the steepest ascent direction for $L_{\text{forget}}$ under the curvature geometry of the retain set, without explicitly enforcing a norm constraint. However, since we have removed the explicit trust region constraint, we need to compensate by adding practical safeguards. We achieve this through step scaling (controlling update magnitudes) and L-BFGS curvature filtering (ensuring numerical stability), as detailed in Appendix \ref{appendix:curvature}.

In practice, we construct $H_{\text{retain}}^{-1}$ using a low-rank L-BFGS \citep{liu1989limited} approximation that incorporates curvature information from both the retain set and a subset of the forget set. This hybrid construction enables the trust region to balance retention of task-specific knowledge with awareness of harmful content boundaries, resulting in more effective influence updates. We discuss implementation details and ablation results in the Appendix \ref{appendix:curvature}.

\subsection{Practical Implementation}

Directly computing and inverting the Hessian matrix $H_{\text{retain}}$ for modern LLMs is computationally intractable due to the enormous parameter space. To address this challenge, we implement two key techniques:

\paragraph{(1) Parameter-Efficient Fine-Tuning.} We apply our method within the Low-Rank Adaptation (LoRA) framework. This reduces the dimensionality of the Hessian matrix to just the trainable parameters, making curvature estimation feasible.

\paragraph{(2) Approximate Hessian Inversion.} We employ L-BFGS (Limited-memory Broyden–Fletcher–Goldfarb–Shanno) to efficiently approximate $H_{\text{retain}}^{-1}$, reducing computation to $\mathcal{O}(mp)$ where $m$ is the memory size and $p$ is the parameter dimensionality. 

This quasi-Newton method builds an approximation of the inverse Hessian through successive low-rank updates, avoiding explicit matrix inversion.

\section{Experimental Results}
\label{sec:experiments}

% Keep the introduction paragraph as is
In this section, we evaluate our curvature-aware alignment restoration approach through empirical analysis across diverse model architectures and safety benchmarks. Our experiments address three fundamental questions: \textbf{(1)} How effectively does our approach restore safety alignment compared to state-of-the-art methods? \textbf{(2)} Can our method restore safety without harming performance or adaptability to new tasks? \textbf{(3)} How robust is the restored alignment against adversarial attacks and parameter perturbations? We first describe our experimental methodology, including architecture selection, fine-tuning protocols, and baseline comparisons. We then present results on safety performance, task utility preservation, in-context learning capabilities, and robustness to both prefilling attacks and weight-space perturbations. Our experimental results demonstrate that the proposed curvature-aware method effectively restores safety alignment while it maintains task performance across diverse model architectures, which addresses a fundamental challenge in LLMs fine-tuning.

\subsection{Experimental Setup}
\label{subsec:exp_setup}

% Keep experimental setup sections as they are
\paragraph{Base LLMs}
We evaluate our curvature-aware alignment restoration method on three representative large language models that span different architectures and training paradigms: LLama-2 7B Chat, LLama 3.1 8B Instruct, and Qwen 2.5 7B Instruct. These models were selected for their widespread adoption in the research community, comparable parameter scales (7-8B parameters), which allow us to assess how our method generalizes across model families with different inherent safety characteristics.

\paragraph{Fine-tuning Protocol}
To maintain computational efficiency while preserving model quality, we implement Parameter-Efficient Fine-Tuning (PEFT) via Low-Rank Adaptation (LoRA). Across all experiments, we utilize a consistent configuration with rank $r=32$ and learning rate $\alpha=2 \times 10^{-4}$. We apply LoRA adapters to the query and value projections in attention layers, following the default configuration used in the PEFT library ~\citep{peft}.

For our primary instruction-tuning dataset, we employ Dolly, a diverse collection of 15,000 human-generated instruction-response pairs spanning multiple domains. We fine-tune each model for $1$ epoch with a batch size of $128$ examples, using the AdamW optimizer. All experiments were conducted on $1$ NVIDIA H100 GPUs with $ 80$ GB memory.

\paragraph{Baseline Methods}
We compare our curvature-aware alignment restoration approach against several state-of-the-art methods for safety-preserving fine-tuning:\textbf{(1) Vanilla Fine-tuning} \citep{hu2022lora}: Standard LoRA fine-tuning without any safety preservation mechanisms, serving as our primary control. \textbf{(2) Vaccine} \citep{huang2024vaccine}: A preventative approach that operates during the initial alignment phase by adding crafted perturbations to hidden embeddings, making the model robust against harmful perturbations that may be introduced during subsequent fine-tuning. \textbf{(3) Safe LoRA} \citep{hsu2024safe}: A data-free, training-free approach that preserves safety alignment during fine-tuning by projecting LoRA weight updates onto an alignment subspace defined by the difference between aligned and unaligned model weights, applying this projection only when updates deviate significantly from the alignment direction. \textbf{(4) SaLoRA} \citep{li2025salora}: A technique that preserves safety alignment during LoRA fine-tuning by introducing a fixed safety module that projects new features to a subspace orthogonal to original safety features, along with task-specific initialization for trainable parameters.

For all baseline methods, we follow the hyperparameter settings recommended in their respective papers, adapting only when necessary to maintain fairness in the comparison.

\subsection{Safety Evaluation}
\label{experiment:safety_eval}

We evaluate model safety on AdvBench, containing 520 adversarial prompts designed to elicit unsafe responses. We allocate 138 samples for constructing the safety matrix required by SaLoRA and reserve the remaining 382 samples for evaluation. Our primary safety metric is the \textit{harmful response rate} (HRR), calculated as the percentage of evaluation samples eliciting unsafe responses. For a comprehensive assessment, we employ both LLama-3 Guard as an automated safety evaluator and human review to validate the quality and accuracy of safety judgments, ensuring a more reliable evaluation of model safety across different methods.

% Place the main results table here for better flow
\begin{table}[t]
\caption{Comparison of safety restoration methods across three model families. HRR (Harmful Response Rate, lower is better) measures safety on AdvBench, while Eval shows performance on fine-tuning dataset (average cross-entropy loss across all examples in the Dolly test set). Utility metrics include four zero-shot tasks: ARC-Challenge (ARC-C), GSM8K, ToxiGen, and TruthfulQA. Our curvature-aware approach achieves best safety across all models while maintaining competitive task performance. Bold indicates best method, underline indicates second-best for each metric within model family.}
\label{tab:main_results}
\centering
\begin{adjustbox}{width=\textwidth}
\begin{tabular}{lccccccc}
\toprule
\multirow{2}{*}{\textbf{Models}} & \multirow{2}{*}{\textbf{Methods}} & \multirow{2}{*}{\textbf{Eval} $\downarrow$} & \multirow{2}{*}{\textbf{HRR} $\downarrow$} & \multicolumn{4}{c}{\textbf{Utility} $\uparrow$} \\
\cmidrule(lr){5-8}
& & & & \textbf{ARC-C} & \textbf{GSM8K} & \textbf{ToxiGen} & \textbf{TruthfulQA} \\
\midrule
\multirow{6}{*}{Llama-3.1 8B} 
& Base & 1.9 & 1.4 & 52.0 & 75.2 & 53.3 & 45.5 \\
\cmidrule(lr){2-8}
& LoRA & \textbf{1.2} & 25.5 & 51.2 & 72.4 & 44.9 & 39.0 \\
& Vaccine & \underline{1.3} & 21.3 & 44.3 & 39.5 & 43.4 & 34.1 \\
& SaLoRA & \textbf{1.2} & \underline{8.1} & \textbf{52.3} & 75.7 & \textbf{49.3} & 41.8 \\
& Safe LoRA & \underline{1.3} & 11.0 & 51.1 & \underline{75.6} & \underline{48.7} & \underline{42.0} \\
& \textbf{Ours} & \underline{1.3} & \textbf{3.0} & \underline{51.8} & \textbf{76.5} & 46.0 & \textbf{43.6} \\
\midrule
\multirow{5}{*}{Qwen 2.5 7B} 
& Base & 3.6 & 0.0 & 53.0 & 76.4 & 57.2 & 56.3 \\
\cmidrule(lr){2-8}
& LoRA & \textbf{1.2} & 24.7 & \textbf{55.0} & 60.2 & \underline{57.2} & 44.5 \\
& Vaccine & \textbf{1.2} & 19.3 & \underline{54.6} & \underline{74.3} & \textbf{57.9} & 44.5 \\
& SaLoRA & \textbf{1.2} & \underline{3.4} & \textbf{55.0} & 69.5 & \underline{57.2} & \underline{49.2} \\
& \textbf{Ours} & \underline{1.4} & \textbf{1.5} & 54.2 & \textbf{75.1} & 57.1 & \textbf{53.3} \\
\midrule
\multirow{5}{*}{Llama-2 7B} 
& Base & 2.5 & 0.0 & 43.3 & 20.1 & 52.9 & 37.2 \\
\cmidrule(lr){2-8}
& LoRA & \textbf{1.1} & 21.4 & 44.4 & 19.6 & 44.7 & 32.3 \\
& Vaccine & \textbf{1.1} & 16.7 & 42.6 & 11.6 & 41.1 & 31.7 \\
& SaLoRA & \textbf{1.1} & \textbf{0.0} & \textbf{45.9} & \textbf{23.6} & \underline{49.5} & \underline{34.7} \\
& Safe LoRA & \underline{1.2} & \textbf{0.0} & \underline{45.6} & 21.5 & 43.8 & 33.1 \\
& \textbf{Ours} & 1.3 & \textbf{0.0} & 44.7 & \underline{22.1} & \textbf{51.7} & \textbf{36.8} \\
\bottomrule
\end{tabular}
\end{adjustbox}
\end{table}

\paragraph{Safety performance}
Table~\ref{tab:main_results} demonstrates our curvature-aware alignment restoration method achieves superior safety results across model families. For Llama-3.1 8B, our approach reduces HRR to just 3.0\%, significantly outperforming both SaLoRA (8.1\%), Vaccine (21.3\%), and Safe LoRA (11.0\%). For Qwen 2.5 7B, we achieve a remarkable 1.5\% HRR, substantially lower than all fine-tuning methods including SaLoRA (3.4\%). With Llama-2 7B, our method successfully restores complete safety alignment (0\% HRR), matching the excellent performance of SaLoRA and Safe LoRA on this model.

\paragraph{Task Performance and Utility Evaluation.}
To show that safety improvements do not compromise task performance, we evaluate models on both the original fine-tuning task (Dolly) and four diverse zero-shot tasks: ARC-Challenge (commonsense reasoning), GSM8K (mathematical reasoning), ToxiGen (toxicity detection), and TruthfulQA (factual consistency). The column `Eval' in Table~\ref{tab:main_results} shows that our method maintains a comparable performance to other safety techniques in the original fine-tuning task, with scores of 1.3, 1.4, and 1.2 in the three model families.

For broader utility, our approach maintains strong performance across tasks. On Llama-3.1 8B, our method achieves the highest scores on GSM8K (76.5) and TruthfulQA (43.6) while maintaining competitive ARC-C performance (51.8). For Qwen 2.5 7B, we obtain the best performance on TruthfulQA (53.3) and GSM8K (75.1). With Llama-2 7B, our approach achieves the highest TruthfulQA (36.8) and ToxiGen (51.7) scores. This demonstrates our curvature-aware method effectively balances safety restoration with preservation of diverse reasoning capabilities.

\subsection{In-Context Learning Performance}
\label{subsec:transfer_ability}

We assess in-context learning capability via few-shot evaluation to determine if alignment restoration preserves the model's adaptability. We measure how different safety restoration methods affect Llama-2 7B's few-shot learning performance across six commonsense reasoning benchmarks. For each task, we compare zero-shot performance with 5-shot performance, where five task examples are included in the prompt before the test instance, allowing the model to perform in-context learning. The improvement from zero-shot to 5-shot performance reflects the model's ability to leverage examples for rapid adaptation, a fundamental capability that should remain intact after safety restoration.

\begin{table}[t]
\caption{In-context learning performance on six commonsense reasoning tasks using Llama-2 7B Chat. Results show 5-shot accuracy percentages with improvements over zero-shot in parentheses. Our curvature-aware method achieves the highest few-shot learning gains on five of six tasks, demonstrating that safety restoration preserves and enhances the model's ability to leverage examples. Bold indicates best absolute performance, while underlines highlight the largest zero-to-five-shot improvements.}
\label{tab:transfer_ability}
\centering
\begin{adjustbox}{width=\textwidth}
\begin{tabular}{lcccccc}
\toprule
\textbf{Methods} & \textbf{ARC-Easy} & \textbf{BoolQ} & \textbf{PIQA} & \textbf{HellaSwag} & \textbf{ARC-Challenge} & \textbf{WinoGrande} \\
\midrule
LoRA & 78.2 (+1.0) & 81.5 (+4.6) & 77.6 (-0.1) & 55.6 (+0.0) & 46.2 (+1.8) & \textbf{72.5} (+3.5) \\
Vaccine & 77.2 (+1.7) & \textbf{82.4} (\underline{+5.0}) & 77.1 (-0.8) & 54.6 (+0.3) & 44.3 (+1.7) & 71.1 (+3.6) \\
SaLoRA & 79.4 (+3.0) & 82.3 (+3.5) & 78.0 (-0.2) & 57.3 (+0.3) & 48.3 (+2.4) & \textbf{72.5} (+3.8) \\
Safe LoRA & 78.9 (+2.6) & 80.7 (+2.2) & 77.8 (-0.7) & 56.5 (+0.0) & 46.8 (+1.2) & 72.2 (+4.1) \\
\textbf{Ours} & \textbf{79.8} (\underline{+4.4}) & 82.2 (+2.5) & \textbf{78.2} (\underline{+1.3}) & \textbf{58.7} (\underline{+0.9}) & \textbf{49.7} (\underline{+5.0}) & 72.2 (\underline{+4.4}) \\
\bottomrule
\end{tabular}
\end{adjustbox}
\end{table}

In Table~\ref{tab:transfer_ability}, our method demonstrates the highest few-shot learning gains on five of six tasks. On ARC-Easy, our approach achieves a substantial +4.4\% improvement over zero-shot, significantly outperforming all baselines, including SaLoRA (+3.0\%) and Safe LoRA (+2.6\%). This pattern continues across other tasks, most notably on ARC-Challenge, where our method achieves a remarkable +5.0\% improvement, more than double that of SaLoRA (+2.4\%).

Notably, our method shows a +1.3\% improvement on PIQA, while all other methods demonstrate minimal or negative transfer. This suggests our curvature-aware approach better preserves the model's commonsense physical reasoning capabilities, which are particularly sensitive to parameter modifications.

% These results show that our curvature-aware restoration not only improves safety alignment but also preserves the model's generalization capabilities. By leveraging geometric structure in the loss landscape, our method recovers broader functionality beyond safety alone. This accounts for the superior transfer performance observed across few-shot tasks.

\subsection{Robustness Evaluation}
\label{subsec:robustness}

We evaluate the robustness of our safety alignment restoration through two distinct experiments: resistance to adversarial prefilling attacks and stability under parameter perturbations.

\subsubsection{Prefilling Attack Resistance}
\label{subsec:prefilling_attacks}

\begin{wrapfigure}{r}{0.5\textwidth}
\centering
\vspace{-12pt}
\includegraphics[width=0.48\textwidth]{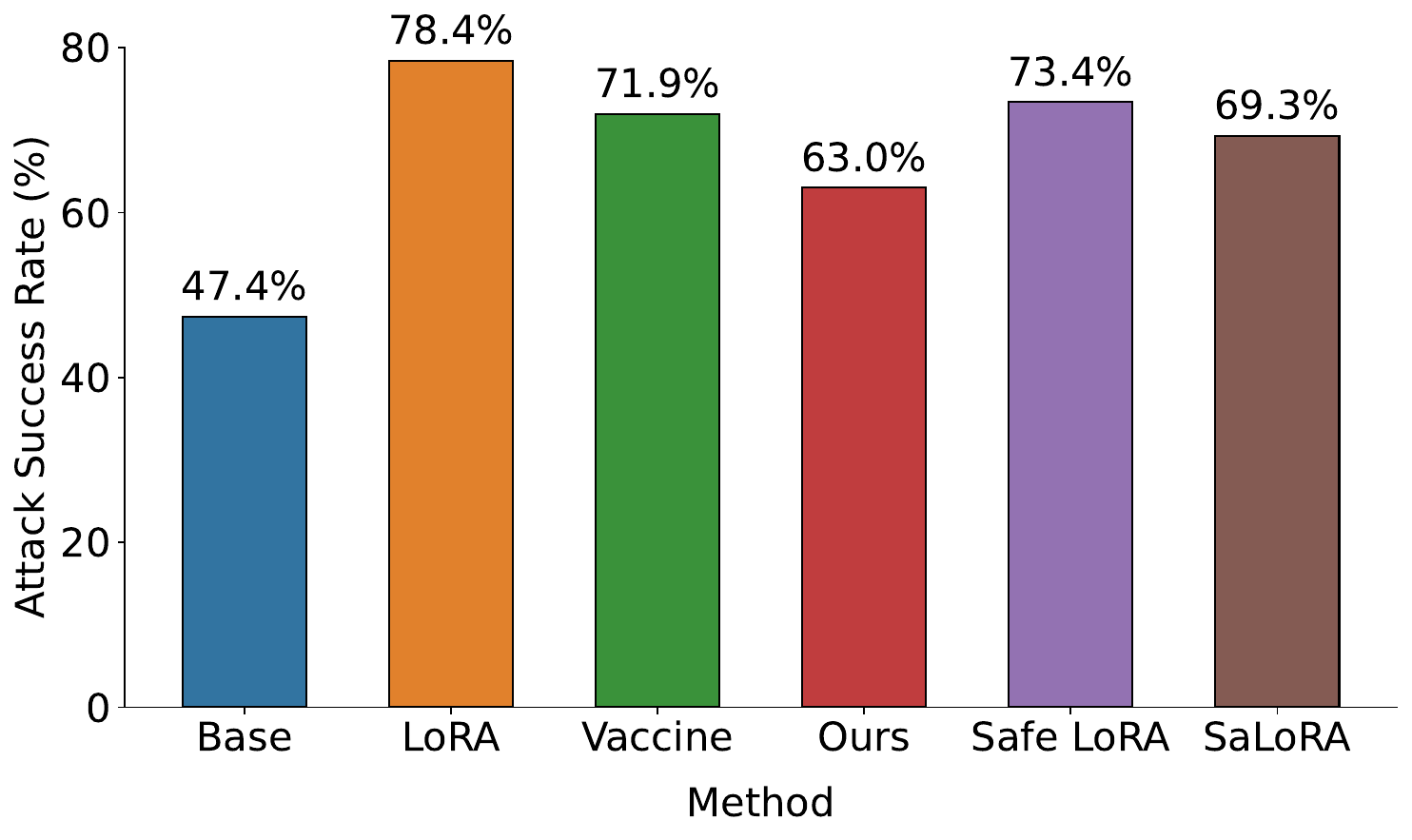}
\caption{Attack success rates (the lower the better) for prefilling attacks across different alignment restoration methods on Llama-3.1 8B evaluated on AdvBench. Our curvature-aware approach achieves 63.0\% ASR, significantly outperforming baseline LoRA (78.4\%) and other safety methods, while approaching the robustness of the base model (47.4\%).}
\label{fig:prefill_attacks}
\vspace{-10pt}
\end{wrapfigure}

This experiment assesses the robustness of our method against inference-time attacks that exploit shallow safety alignment vulnerabilities in LLMs~\citep{qi2024safety}.

\paragraph{Experimental Setup}  
We use 382 adversarial prompts from AdvBench (used in Section~\ref{experiment:safety_eval}) to simulate a prefilling attack. Following prior work~\citep{qi2024safety, andriushchenko2024jailbreaking}, each input is prepended with four non-refusal tokens, which are designed to bypass the model's standard safety refusal mechanisms.\footnote{Details of the non-refusal token construction are provided in Appendix~\ref{appendix:prefill}.}

We evaluate models fine-tuned with five different methods: vanilla LoRA, Vaccine~\citep{huang2024vaccine}, Safe LoRA~\citep{hsu2024safe}, SaLoRA~\citep{li2025salora}, and our curvature-aware approach. We report \textbf{attack success rate (ASR)} as the percentage of inputs that lead to harmful completions (lower is better).

\paragraph{Results Analysis}  
As shown in Figure~\ref{fig:prefill_attacks}, our method achieves a lower ASR (63.0\%) than all other alignment restoration baselines. Compared to standard LoRA fine-tuning (78.4\%), our approach yields a 19.6\% relative reduction in attack success, and also demonstrates improved robustness over Vaccine, Safe LoRA, and SaLoRA. These findings highlight the effectiveness of our curvature-aware approach in mitigating shallow alignment vulnerabilities and preserving safety under adversarial prompting.

\subsubsection{Parameter Perturbation Stability}
\label{subsec:safety_basin}

\begin{wraptable}{r}{0.4\textwidth}
\vspace{-12pt}
\centering
\caption{VISAGE scores measuring safety basin robustness. Higher scores indicate more robust safety basins resistant to parameter perturbations. Our approach achieves 56.1, substantially outperforming all baselines.}
\label{tab:visage_scores}
\begin{tabular}{lc}
\toprule
\textbf{Method} & \textbf{VISAGE Score} \\
\midrule
LoRA & 21.1 \\
Vaccine & 28.8 \\
SaLoRA & 32.1 \\
\textbf{Ours} & \textbf{56.1} \\
\bottomrule
\end{tabular}
\vspace{-10pt}
\end{wraptable}

We further evaluate the robustness of alignment restoration methods under parameter perturbations by analyzing the \emph{safety basin} ~\cite{peng2024navigating}, which refers to the region in parameter space where the model continues to behave safely despite small changes.

\paragraph{Experimental Setup}  
We test the Qwen 2.5 7B Instruct model fine-tuned with four methods: vanilla LoRA (as the baseline), and three safety alignment techniques: Vaccine~\citep{huang2024vaccine}, SaLoRA~\citep{li2025salora}, and our curvature-aware approach. For each model, we apply parameter perturbations along randomly sampled directions, varying the perturbation magnitude within the range \([-0.5, 0.5]\).

\begin{wrapfigure}{r}{0.5\textwidth}
\vspace{-12pt}
\centering
\includegraphics[width=0.48\textwidth]{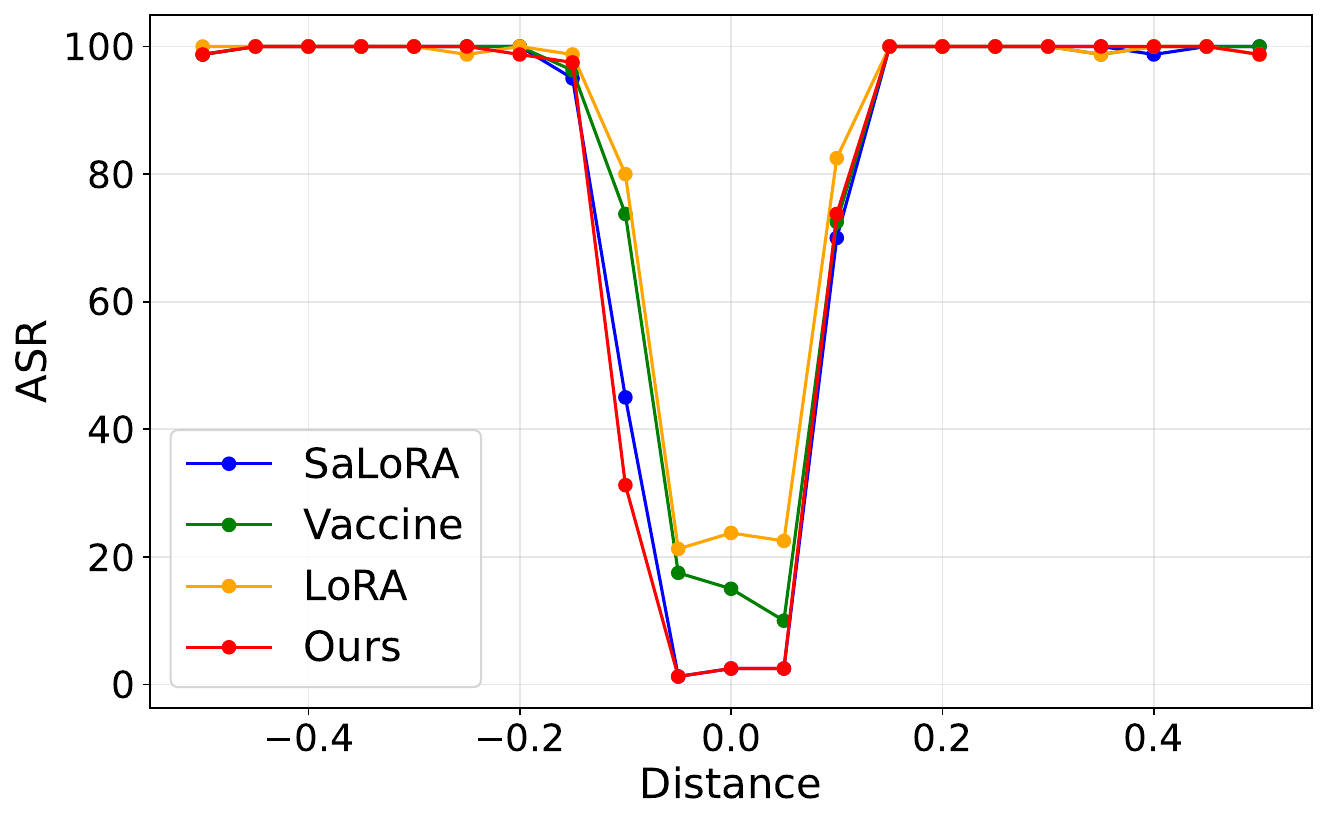}
\caption{Safety landscape visualization showing Attack Success Rate (ASR) across parameter perturbations for different methods on Qwen 2.5 7B. Our approach maintains a significantly wider and deeper safety basin, with near 0\% ASR at the origin and slower degradation with distance.}
\label{fig:safety_basin}
\vspace{-10pt}
\end{wrapfigure}

We compare the \textbf{attack success rate (ASR)} at each perturbation level and compute the VISAGE score~\citep{peng2024navigating}, which measures the average safety margin across all directions. A higher VISAGE score indicates that the model remains safe under a wider range of parameter variations.

\paragraph{Results Analysis}  
As shown in Table~\ref{tab:visage_scores} and Figure~\ref{fig:safety_basin}, our curvature-aware method achieves the highest VISAGE score (56.1), substantially outperforming SaLoRA (32.1), Vaccine (28.8), and LoRA (21.1). The safety landscape visualization confirms that our method maintains a broader and deeper safety basin, with nearly zero ASR at the origin and slower degradation as perturbation magnitude increases. These results indicate that our method produces more resilient safety alignment, offering stronger robustness to parameter noise and adaptation.
\section{Conclusion}

We present a curvature-aware alignment restoration framework that addresses the challenge of safety degradation in fine-tuned LLMs. Our approach builds on the empirical observation that the loss landscape associated with harmful content remains structurally preserved after task-specific fine-tuning. Leveraging this geometric insight, we apply influence functions and second-order optimization to selectively increase loss on harmful inputs while maintaining task performance. Extensive evaluations across multiple model families and adversarial settings demonstrate that our method consistently reduces harmful responses while preserving few-shot generalization and utility on downstream tasks.

\paragraph{Limitations and Future Work}
Our implementation focuses on LoRA parameters, simplifying curvature estimation at the cost of expressiveness for deep interventions. Effectiveness may vary in settings with entangled safety and task objectives (e.g., creative content generation), despite strong performance in our experiments. Future work could extend to full-model fine-tuning with scalable curvature approximations or incorporate adversarial training for enhanced robustness in high-risk scenarios.

\bibliographystyle{unsrtnat}
\bibliography{references}
\appendix
\clearpage
\appendix
\section{Related Work}

\subsection{Safety and Robustness in Large Language Models}

\paragraph{Safety Alignment in Large Language Models}
Ensuring the safety of large language models (LLMs) has become a central research challenge as their deployment expands into high-stakes domains. Models pretrained on vast internet corpora often internalize harmful behaviors, prompting the development of post-training alignment methods such as Reinforcement Learning from Human Feedback (RLHF)~\citep{ouyang2022training,dai2023safe} and supervised instruction tuning~\citep{bai2023constitutional,zhang2023instruction,zhou2023lima}. Despite their effectiveness, these safety mechanisms remain fragile, with studies showing that fine-tuning aligned models on downstream tasks can lead to significant safety degradation~\citep{qi2023safety,qi2023finetuning}. Concurrently, \textit{parameter-efficient fine-tuning} (PEFT) techniques have emerged to adapt large models with minimal updates. Low-Rank Adaptation (LoRA)~\citep{hu2021lora} has become particularly popular by constraining updates to low-rank matrices applied to the model's weight matrices, significantly reducing trainable parameters while maintaining performance. Building on LoRA's efficiency, several \textit{safety-preserving fine-tuning} approaches have been developed to address safety degradation. Vaccine~\citep{huang2024vaccine} introduces adversarial perturbations during training to immunize models against unsafe queries. SafeLoRA~\citep{hsu2024safe} extends LoRA by projecting weight updates onto an alignment subspace defined by the difference between aligned and unaligned model weights. Similarly, SaLoRA~\citep{li2025salora} preserves safety during LoRA fine-tuning by introducing a fixed safety module that projects new features to a subspace orthogonal to original safety features. However, these approaches typically either compromise task performance or rely on heuristic projections without geometric insights. Recent findings suggest that safety-relevant behaviors occupy distinct, resilient regions in the loss landscape~\citep{peng2024visage}, indicating that geometric properties of the parameter space could enable more robust alignment preservation~\citep{li2025safety,arditi2024refusal}. Our work builds on these geometric insights by employing influence functions and curvature-aware optimization to restore safety alignment without sacrificing task performance. Unlike previous approaches that use heuristic constraints, our method directly leverages the preserved structure of the loss landscape to navigate toward parameter configurations that enhance safety while maintaining model capabilities.

\subsection{Unlearning and Parameter Space Geometry}

When harmful behaviors emerge in LLMs following fine-tuning, \textit{machine unlearning} offers a principled framework to selectively remove them ~\citep{huu2024effects,li2024wmdp,liu2024large}. Influence function-based unlearning~\citep{koh2017understanding, chen2023comprehensive,yuan2024closer} estimates the gradient direction that increases the loss on undesired examples while minimally impacting desired behaviors, effectively reversing their influence in parameter space ~\citep{liu2025rethinking,barez2025open}. Other approaches such as SISA~\citep{bourtoule2021machine} or trust-region unlearning~\citep{golatkar2020forgetting} offer certified deletion by retraining from strategically partitioned checkpoints. However, these methods often incur high computational costs or suffer from degraded generalization. In parallel, \textit{curvature-aware optimization} techniques have been explored to control model drift during fine-tuning. Elastic Weight Consolidation~\citep{kirkpatrick2017overcoming} and similar continual learning strategies use curvature estimates (e.g., Fisher information) to constrain updates in directions that preserve previously acquired capabilities. Trust-region policy optimization~\citep{schulman2015trust} and natural gradient methods~\citep{amari1998natural} apply second-order constraints to keep parameter updates within functionally safe neighborhoods. Our method unifies these perspectives by framing safety restoration as a second-order constrained optimization problem over the loss landscape. We employ influence functions and L-BFGS-based curvature estimation to direct updates that increase loss on harmful content while staying within a trust region defined by the retain set, enabling scalable and stable safety restoration in fine-tuned LLMs.
\section{Detailed Implementation}
\subsection{Curvature-Aware L-BFGS Construction}
\label{appendix:curvature}
\paragraph{Data Partitioning.}
To avoid overlap between curvature estimation and influence-based safety restoration, we partition the HEx-PHI dataset~\citep{qi2023finetuning}, which contains 330 adversarially constructed harmful prompts. For L-BFGS curvature pair construction, we use a total of 256 examples (64 examples from each of four batches) selected from HEx-PHI as part of the mixed curvature set \(D^{\text{curv}}_{\text{forget}}\). Separately, to compute the forget loss \(\mathcal{L}_{\text{forget}}\), we reserve 50 held-out examples from the remaining HEx-PHI data (named as $\mathcal{D}_{\text{forget}}$). These examples are not used during curvature approximation and are exclusively employed to evaluate or guide updates that suppress harmful generations. This partitioning ensures clean separation between curvature modeling and influence-based optimization targets.

To approximate the inverse Hessian $H^{-1}_{\text{retain}}$ in Equation \ref{eq:influence_max}, we construct a low-rank L-BFGS history over LoRA parameters using a carefully designed curvature buffer. This buffer integrates information from three strategically selected disjoint datasets: a subset of the forget set $\mathcal{D}_{\text{forget}}^{\text{curv}}$ (HEx-PHI dataset) and two distinct subsets of the retain set $\mathcal{D}_{\text{retain}}^{(1)}, \mathcal{D}_{\text{retain}}^{(2)}$ (derived from the fine-tuning dataset). This multi-dataset approach ensures the captured curvature spans both safety-critical and task-aligned directions in parameter space.

Each L-BFGS pair $(s_t, y_t)$ is computed via gradient accumulation over batches of 64 examples. Our empirical analysis reveals that just 10 high-quality pairs sufficiently approximate the local curvature structure for effective influence updates. We allocate these pairs approximately equally across the three datasets, requiring a minimum of 192 examples per set. To enhance curvature diversity, we employ varying learning rates (0.001, 0.002, 0.005) across optimization steps. A trust region $\delta_t$ constrains update magnitudes by scaling steps to a bounded norm, while a reduction ratio $\rho_t$ determines step acceptance and dynamically adjusts $\delta_t$.

To ensure robust and numerically stable curvature estimation, we implement several filtering mechanisms: \textbf{(1)} rejecting curvature pairs with insufficient curvature ($\langle s_t, y_t \rangle < 10^{-6}$), \textbf{(2)} normalizing $s_t, y_t$ vectors to unit norm before storage, \textbf{(3)} applying adaptive damping when negative curvature is encountered, and \textbf{(4)} excluding pairs with degenerate step or gradient norms. These safeguards collectively prevent ill-conditioning in the inverse Hessian approximation.

In practice, we recompute curvature pairs at the beginning of each safety restoration iteration. Our experiments demonstrate that just three such iterations suffice for effective alignment restoration across all evaluated model architectures, and this 3-step procedure is consistently employed throughout our experimental validation.

% \paragraph{Pseudocode: Trust-Region L-BFGS History Builder}

\begin{algorithm}[H]
\caption{Curvature-Aware L-BFGS History Construction}
\begin{algorithmic}[1]
\State \textbf{Input:} Model $f_\theta$, datasets $\mathcal{D}_{\text{forget}}^{\text{curv}}, \mathcal{D}_{\text{retain}}^{(1,2)}$, LoRA parameters $\theta_{\text{LoRA}}$
\State Initialize empty history lists: $\mathcal{S}, \mathcal{Y}$
\State Set initial trust radius $\delta = 0.05$
\For{$t = 1$ to $T$}
    \State Sample batch $B_t$ from one of the datasets (round-robin)
    \State Compute initial loss $\mathcal{L}_{\text{init}}$ and gradients $g_t$
    \State Propose step $d_t = -g_t$ and rescale to $\|d_t\| \leq \delta$
    \State Save $\theta_t$, apply step to get $\theta_{t+1}$
    \State Compute final loss $\mathcal{L}_{\text{final}}$ and gradients $g_{t+1}$
    \State Compute actual and predicted reduction, ratio $\rho_t$
    \If{$\rho_t < 0.25$} 
        \State Shrink trust radius: $\delta \gets 0.5 \delta$
        \State Revert to $\theta_t$
        \State \textbf{continue}
    \ElsIf{$\rho_t > 0.75$}
        \State Expand trust radius: $\delta \gets 1.5 \delta$
    \EndIf
    \State Compute $s_t = \theta_{t+1} - \theta_t$, $y_t = g_{t+1} - g_t$
    \If{$\langle s_t, y_t \rangle > \epsilon$}
        \State Normalize $s_t$, $y_t$, add to $\mathcal{S}, \mathcal{Y}$
    \EndIf
\EndFor
\State \Return $\mathcal{S}, \mathcal{Y}$
\end{algorithmic}
\end{algorithm}

\subsection{Influence Update Mechanism}
\label{appendix:loss}

To prevent overcorrection and preserve generalization capabilities of the model during alignment restoration, we apply L2 regularization to the influence-based update direction $\Delta \theta$. At each iteration, the update to the LoRA parameters is computed as:

\[
\theta_{\text{new}} = \theta_{\text{tuned}} + \eta \cdot \Delta \theta - \lambda \cdot \theta,
\]

where:
\begin{itemize}
    \item $\eta$ is the update scale (determined by a fixed multiplier or small grid search),
    \item $\Delta \theta$ is the L-BFGS-projected gradient direction (from Appendix~\ref{appendix:curvature}),
    \item $\lambda$ is the L2 regularization weight, progressively annealed across iterations (e.g., $\lambda \leftarrow 0.95 \cdot \lambda$).
\end{itemize}

% This L2 penalty acts as a quadratic anchor that resists large deviations from the original fine-tuned weights, helping to preserve retained task performance. Compared to unconstrained influence-based updates, this leads to improved safety-utility trade-offs and prevents drift toward harmful or unstable regions of the parameter space.

\paragraph{Unlearning Objective}The harmful gradient $\nabla \mathcal{L}_{\text{forget}}$ is obtained by evaluating the model on the forget set using a cross-entropy loss:

\[
\mathcal{L}_{\text{forget}} = \text{CE}(\hat{y}, y)
\]

% \subsection{Pseudocode: L2-Regularized Influence Update}
\begin{algorithm}[H]
\caption{Safety Restoration via Influence Update}
\begin{algorithmic}[1]
\State \textbf{Input:} LoRA parameters $\theta$, L-BFGS history $(\mathcal{S}, \mathcal{Y})$, forget dataset $\mathcal{D}_{\text{forget}}$, step size $\eta$, L2 weight $\lambda$
\State Initialize accumulated gradient $g \gets 0$
\For{each batch $B$ in $\mathcal{D}_{\text{forget}}$}
    \State Compute loss $\mathcal{L}_{\text{forget}} = \text{CE}(f_\theta(B))$
    \State Compute gradient $\nabla \mathcal{L}_{\text{forget}}$ and accumulate into $g$
\EndFor
\State Project $g$ through inverse Hessian: $\Delta \theta \gets -H^{-1} g$ using L-BFGS (see Appendix~\ref{appendix:curvature})
\State Normalize $\Delta \theta \gets \Delta \theta / \|\Delta \theta\|$
\State \textbf{for} each parameter $\theta_i$ in LoRA:
\State \quad Extract corresponding slice $\Delta \theta_i$
\State \quad Compute L2-regularized update:
\[
\theta_i \gets \theta_i + \eta \cdot \Delta \theta_i - \lambda \cdot \theta_i
\]
\State \textbf{return} Updated parameters $\theta$
\end{algorithmic}
\end{algorithm}
\subsection{Loss Landscape Visualization Implementation}
\label{appendix:loss_visualize}
This section provides a detailed description of our methodology for visualizing the loss landscapes of language models before and after fine-tuning.

\paragraph{Gradient-Informed Direction Generation}
Unlike conventional approaches that use random directions in parameter space, we generate perturbation directions informed by gradients computed on the model's loss function. For computational tractability, we focus only on attention and MLP layers, which most strongly influence model behavior. For each perturbation direction $\mathbf{d}_i$, we calculate:
\begin{align}
\mathbf{d}_i = \text{RandomScale}(\nabla_\theta \mathcal{L}(\theta))
\end{align}
where $\nabla_\theta \mathcal{L}(\theta)$ represents accumulated gradients from a fixed set of validation samples, and $\text{RandomScale}(\cdot)$ applies random scaling factors to different parameters using a direction-specific random seed. We generate two perturbation directions $\mathbf{d}_1$ and $\mathbf{d}_2$ using different random seeds (1000 and 2000), which affects the scaling factors applied to the gradients. Due to the high dimensionality of the parameter space, these two directions are approximately orthogonal with high probability.

\paragraph{Grid Construction and Evaluation}
To visualize the loss landscape, we construct a 2D grid in parameter space by varying perturbation magnitudes along these two directions:
\begin{align}
\theta_{i,j} = \theta_\text{original} + \lambda_i \cdot \mathbf{d}_1 + \lambda_j \cdot \mathbf{d}_2
\end{align}
where $\lambda_i, \lambda_j \in [-\alpha, \alpha]$ are scalar coefficients with $\alpha = 0.01$. We construct a $20 \times 20$ grid by uniformly sampling $\lambda$ values. For each grid point $\theta_{i,j}$, we compute the model's loss on both harmful and benign datasets, creating separate loss landscapes for each model state.

\paragraph{Memory-Optimized Implementation}

Large language models present significant memory challenges for loss landscape visualization. To address this, we implement several optimizations: row-by-row processing to compute one grid row at a time; parameter subsetting that applies perturbations only to attention and MLP layers; gradient accumulation over small batches; and bfloat16 precision for all computations. These techniques allow us to visualize loss landscapes of multi-billion parameter models without excessive memory requirements.

\paragraph{Structural Difference Quantification}

To quantify the structural similarity between base and fine-tuned model loss landscapes, we define a correlation-based metric:

\begin{align}
\text{StructDiff} = (1 - |\text{corr}(\nabla^2\mathcal{L}_\text{base}, \nabla^2\mathcal{L}_\text{tuned})|) \times 100\%
\end{align}

where $\nabla^2\mathcal{L}$ is the Laplacian of the loss landscape, approximated using finite differences on our computed grid. This metric captures differences in curvature patterns rather than absolute loss values, providing a more reliable measure of structural preservation. The figure visualizes a visualization on LLama-3-8B, under LoRA fine-tuning (where hyperparameters are used the same as in the main experiments), which further confirms the model preserver structure on harmful data compared to the benign ones.
\begin{figure}[h]
\centering
\includegraphics[width=\textwidth]{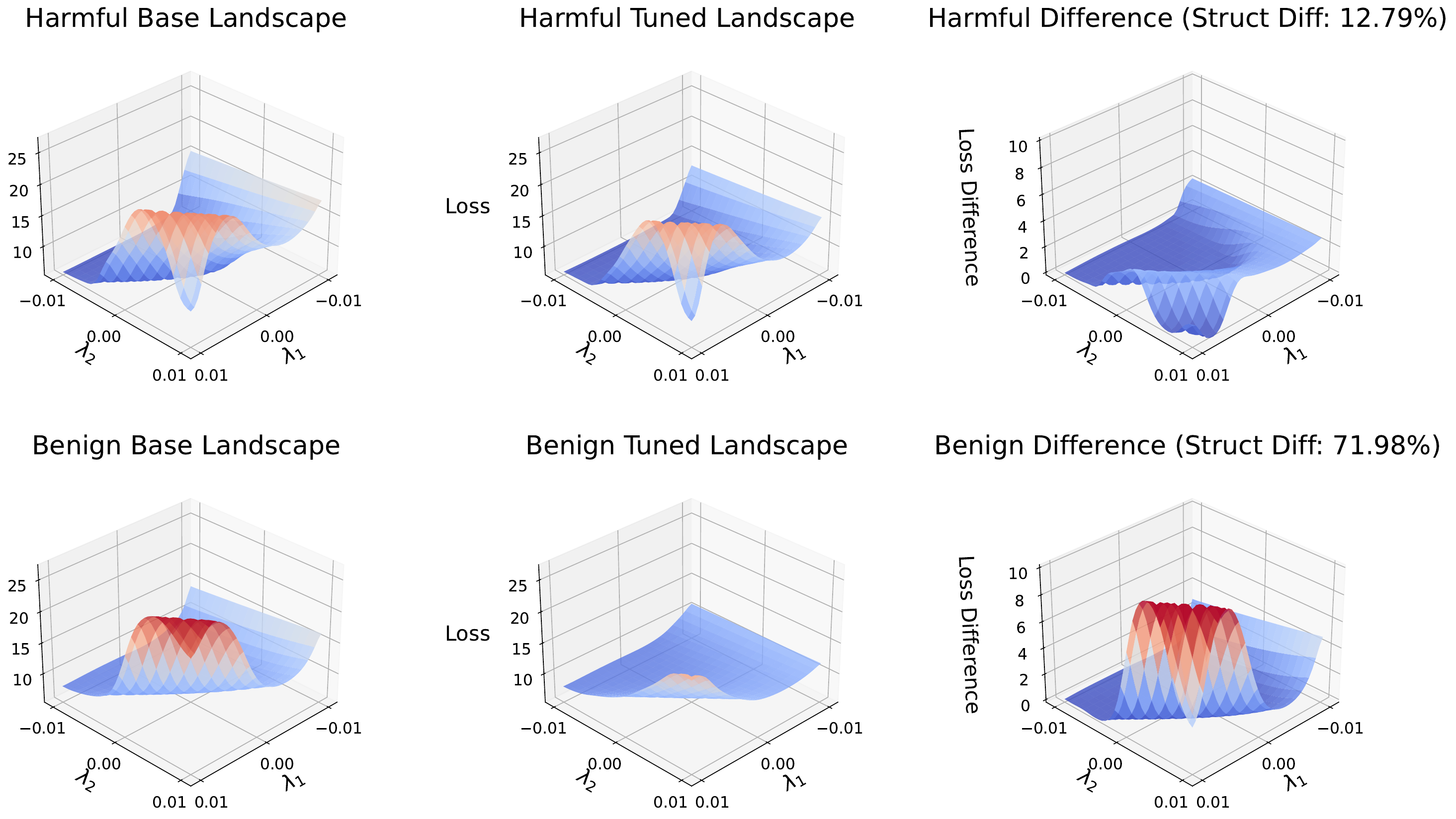}
\caption{3D loss landscape visualization for LLaMA-3 8B with LoRA fine-tuning using gradient-informed direction projection. Top row: harmful content (HEx-PHI); bottom row: general data (Alpaca). LoRA fine-tuning preserves the loss landscape structure for harmful content (12.79\% structural difference) while substantially altering general data landscapes (71.98\% structural difference), demonstrating that parameter-efficient methods similarly maintain safety-relevant geometric features.}
\label{fig:loss_landscape_lora}
\end{figure}
\paragraph{Cross-Sectional Analysis}

\begin{figure}[t]
    \centering
    \includegraphics[width=\linewidth]{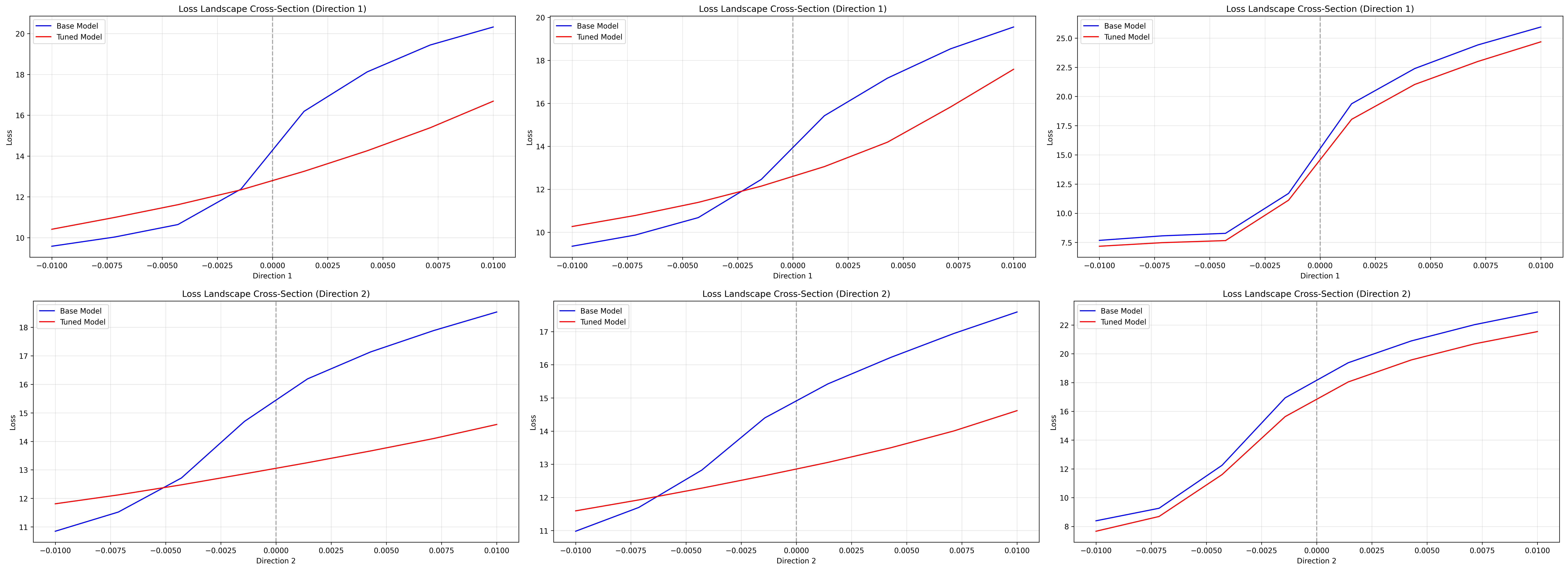}
    \caption{Loss landscape cross-sections along two perturbation directions for base (blue) and fine-tuned (red) models utilizing Qwen 2.7 7B Instruct across three datasets: Dolly (left), Alpaca (middle), and HEx-PHI harmful content (right). While task-specific and general datasets show significant divergence between models, harmful content exhibits remarkable structural similarity with preserved curvature characteristics, particularly near the origin (0,0).}
    \label{fig:cross_loss}
\end{figure}

To provide additional insight into the loss landscape structure, we extract cross-sectional views along each perturbation direction at the origin point. Figure~\ref{fig:cross_loss} shows these cross-sections across our three evaluation datasets. For Dolly and Alpaca datasets (left and middle columns), we observe significant structural divergence between base and fine-tuned models. The fine-tuned model consistently exhibits lower loss values in negative direction regions, reflecting optimization for task-specific objectives. The intersection points where the curves cross represent transition zones in parameter space where model behaviors begin to diverge more dramatically.

On the other hand, for harmful content (right column), the base and fine-tuned model loss curves remain remarkably parallel with nearly identical structural features. Both models show similar sharp increases in loss (forming "cliff" patterns) at similar positions along both directions. This preserved geometric correspondence provides quantitative evidence for our hypothesis that safety-relevant regions in the loss landscape maintain their structural integrity during fine-tuning. These cross-sectional visualizations complement our 3D surface plots and structural difference metrics, providing a more granular view of how loss landscapes change along specific directions of interest.
\subsection{Prefill Attack Construction and Non-Refusal Token Injection}
\label{appendix:prefill}
To simulate prefilling attacks, we adopt the setup introduced in AdvBench~\citep{zou2023universal}, which provides a collection of adversarial goal-target pairs designed to bypass refusal mechanisms in safety-aligned language models. Each goal represents a harmful instruction, and the corresponding target is a benign-looking prefix that avoids immediate refusal while steering the model toward unsafe completions. In our setup, we construct the prefilled input by first applying a prompt template to each goal, then appending the associated target prefix directly to the end of the prompt. The resulting input is passed to the model, forcing it to generate from a context that includes several non-refusal tokens. We use a fixed number of prefix tokens (e.g., the first 4 tokens from each target) to ensure consistent perturbation across examples. This approach effectively bypasses shallow safety filters by shifting the harmful intent away from the beginning of the prompt, thereby exposing vulnerabilities in the model’s alignment mechanisms.

% \subsection{Visualization Techniques}

% For visual clarity, we generate multiple complementary visualizations: 3D surface plots showing the complete loss landscape for both models; contour plots highlighting level sets; difference plots showing the point-wise subtraction of landscapes; and cross-section plots along each perturbation direction. All visualizations use consistent color schemes and scaling to facilitate direct comparison between base and fine-tuned model landscapes.
\section{Ablation Study}
\subsection{Recovery of Base Model Behavior}
\begin{figure}[t]
    \centering
    \includegraphics[width=0.95\linewidth]{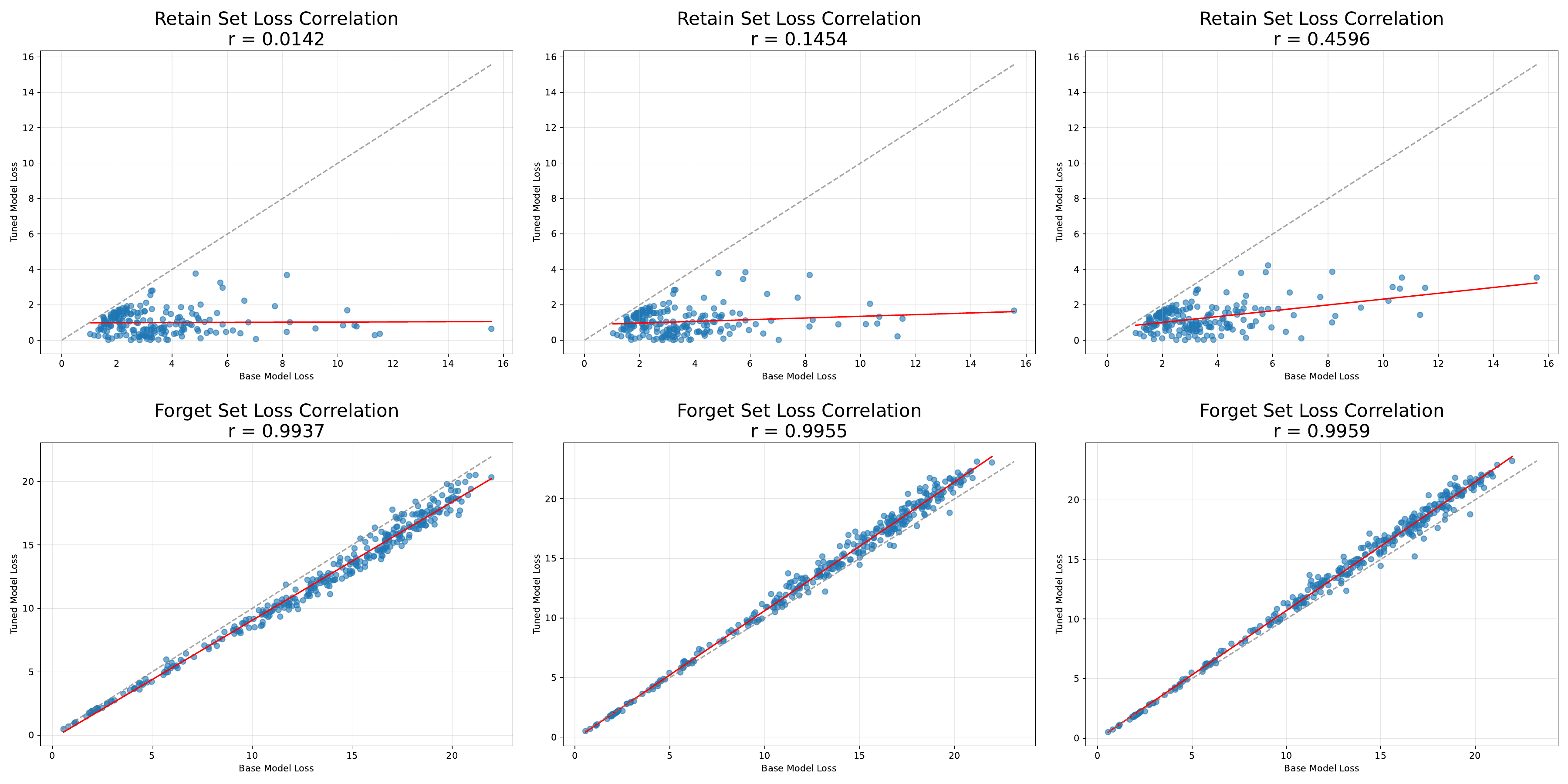}
    \caption{Correlation analysis during restoration. Top row: Dolly (test set) correlation improves from $r=0.014$ to $r=0.456$, showing functional recovery. Bottom row: Forget set correlation stabilizes at $r=0.996$, demonstrating realignment with base model behavior in harmful regions.}
    \label{fig:loss_correlation}
\end{figure}
A key result of our alignment restoration approach is its ability to recover the original base model's safety behavior patterns. To verify this property, we analyze the relationship between the restored model and the base model throughout the recovery process on Qwen 2.5 7B Instruct model. Figure~\ref{fig:loss_correlation} illustrates Pearson correlation coefficients between the base model and the restored model on a held-out evaluation set (Dolly) under increasing restoration steps. We report correlations on both the retain (Dolly) set and forget (harmful) set, computed between per-example loss values as a proxy for functional alignment.

Initially, the fine-tuned (unsafe) model exhibits near-zero correlation with the base model on the retain set (e.g., $r = 0.014$), indicating severe deviation. As alignment restoration progresses, the correlation increases steadily (e.g., $r = 0.145$, then $r = 0.460$), reflecting functional recovery. On the forget set, we observe near-perfect preservation of the base model's loss ranking by the third step ($r = 0.996$), suggesting that the restored model re-aligns closely with the base behavior in harmful regions. These results support the hypothesis that the safety properties of the base model remain geometrically accessible even after fine-tuning, and that our method effectively re-navigates the loss landscape to recover them.

% \subsection{Connection Between Ours And SAM (Sharpness-aware Minimization)}
% \subsection{The stability of L-BFGS trust region}
\subsection{Comparison with First-Order Methods}
\begin{figure}[t]
\centering
\includegraphics[width=0.95\textwidth]{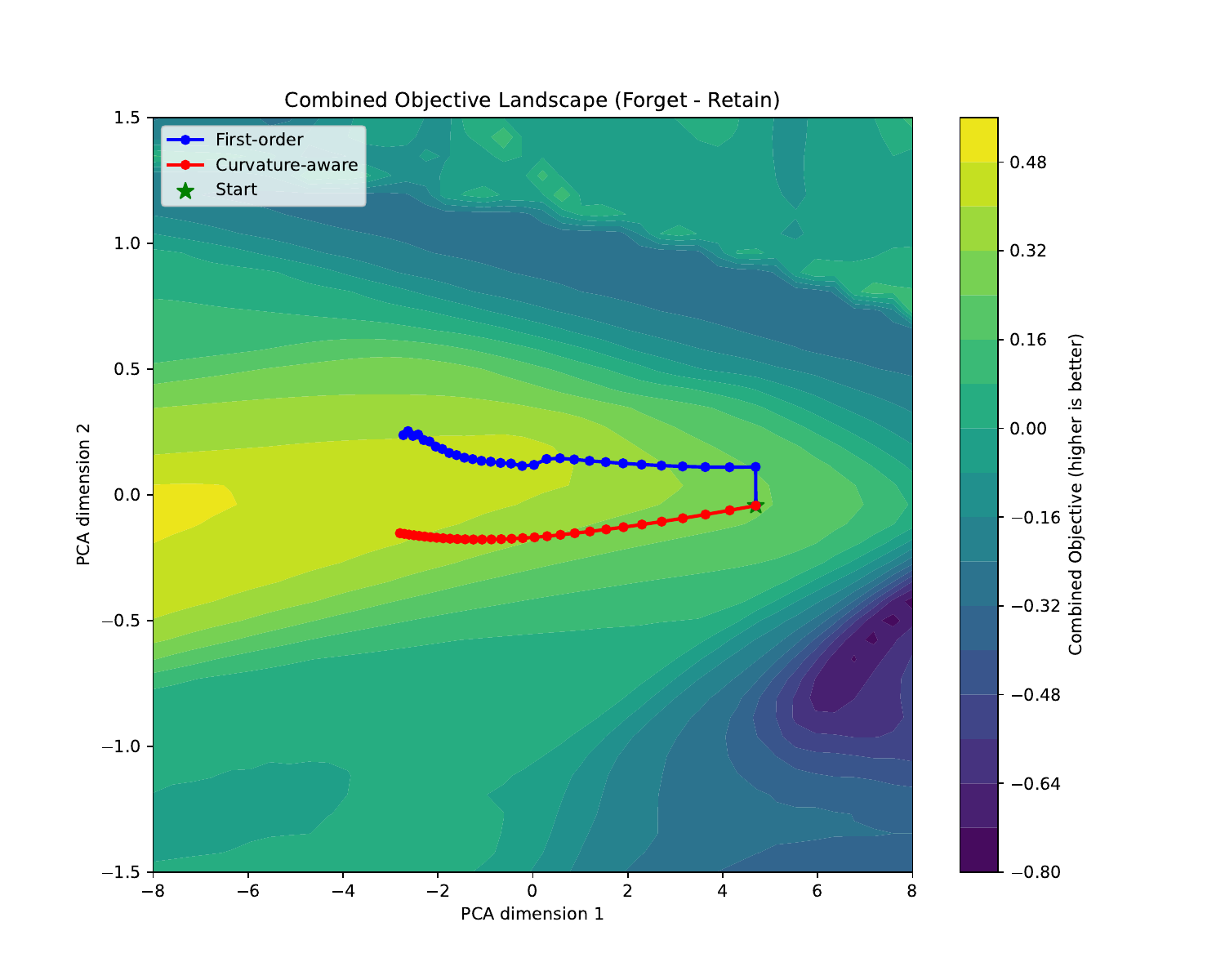}
\caption{Parameter space navigation comparison between first-order (blue) and curvature-aware (red) methods at a conservative learning rate (0.01), projected onto the first two principal components. The contour plot shows the combined objective landscape (forget loss minus retain loss), where higher values (yellow) represent more effective safety restoration while preserving task performance. Our curvature-aware approach follows a more direct path through higher-value regions, demonstrating superior landscape navigation. Both methods start from the same fine-tuned model parameters (green star).}
\label{fig:optimization_comparison_low_lr}
\end{figure}
\begin{figure}[t]
\centering
\includegraphics[width=0.95\textwidth]{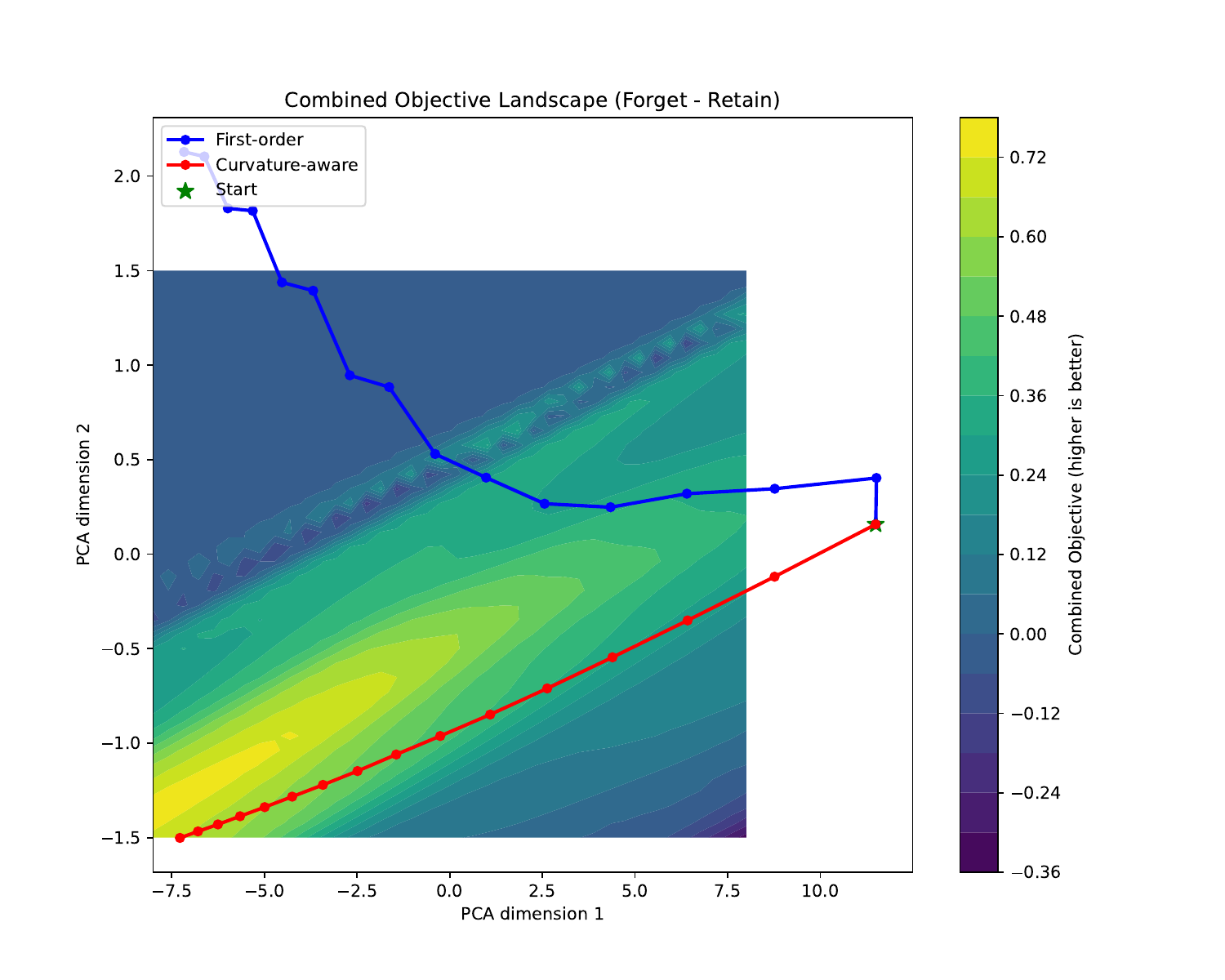}
\caption{Parameter space navigation comparison at a higher learning rate (0.05). The first-order method (blue) exhibits extreme oscillation and instability, making large erratic movements and repeatedly venturing into negative-value regions (purple/dark blue). In contrast, our curvature-aware method (red) demonstrates remarkable stability, following an almost perfectly straight path that steadily progresses through higher-value regions. This visualization highlights how curvature awareness provides robustness to hyperparameter choices and avoids wasteful exploration of the parameter space.}
\label{fig:optimization_comparison_high_lr}
\end{figure}

First-order optimization methods dominate machine unlearning approaches due to their computational efficiency. However, these methods struggle with the complex non-convex landscapes characteristic of fine-tuned LLMs. Our curvature-aware approach fundamentally improves upon first-order methods by incorporating second-order information about the loss landscape's geometry. Figures~\ref{fig:optimization_comparison_low_lr} and~\ref{fig:optimization_comparison_high_lr} visualize the optimization trajectories of our curvature-aware method versus a representative first-order approach at different learning rates. We project the high-dimensional parameter space into a 2D representation using Principal Component Analysis (PCA) on parameter updates during optimization. The contour plots represent the combined objective landscape, where higher values (yellow regions) indicate better safety restoration while preserving task performance.

At a conservative learning rate (0.01, Figure~\ref{fig:optimization_comparison_low_lr}), the first-order method (blue trajectory) exhibits inefficient navigation, following a suboptimal path that initially makes progress but then traverses through lower-value regions. In contrast, our curvature-aware approach (red trajectory) identifies and follows a more direct path toward the high-value region, demonstrating superior awareness of the landscape's geometry. At a higher learning rate (0.05, Figure~\ref{fig:optimization_comparison_high_lr}), the limitations of first-order methods become even more pronounced. The blue trajectory exhibits dramatic oscillations and instability, making large, erratic movements through parameter space. Our curvature-aware method maintains remarkable stability even at this higher learning rate, following an almost perfectly straight path that steadily progresses through increasingly favorable regions of the objective landscape.

\subsection{Connection to Machine Unlearning}
\label{appendix:unlearning}
\begin{figure}[t]
\centering
\includegraphics[width=\textwidth]{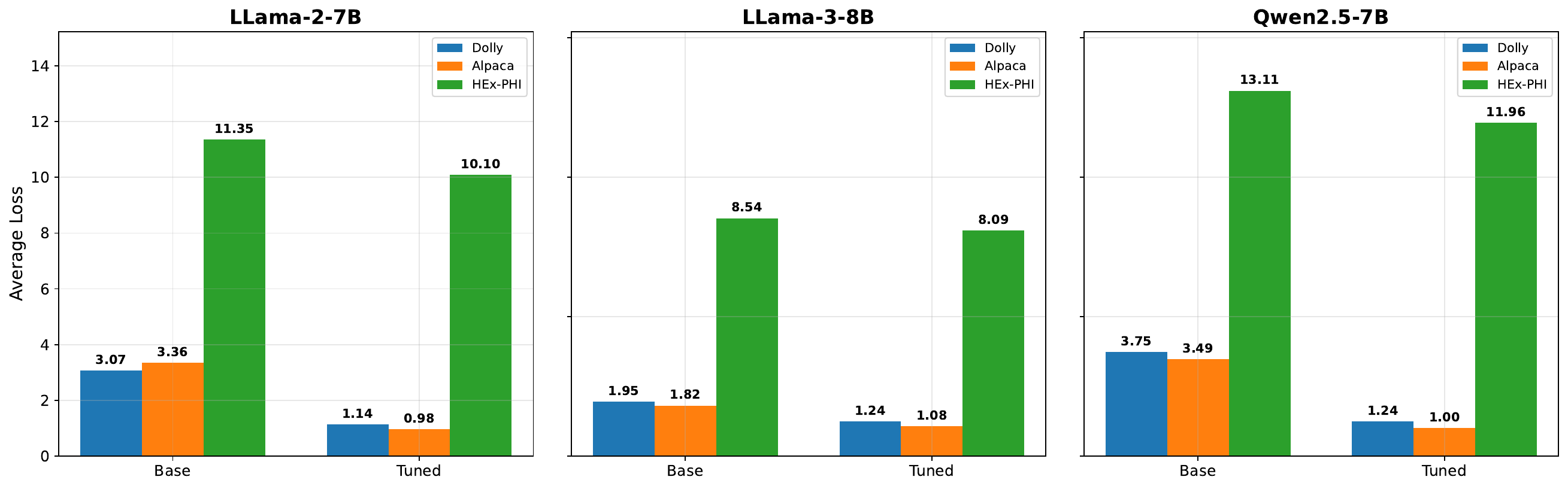}
\caption{Average loss comparison across base and fine-tuned models for three datasets: Dolly (task-specific), Alpaca (general), and HEx-PHI (harmful). Across all three model families, harmful content consistently exhibits substantially higher loss (8.09-13.11) compared to benign content (0.98-3.75) in base models. Fine-tuning reduces loss on both task-specific and general content while simultaneously reducing the loss gap on harmful content.}
\label{fig:loss_comparison}
\end{figure}

Our curvature-aware alignment restoration framework connects to machine unlearning by effectively reversing the unintended "forgetting" of safety behaviors that occurs during fine-tuning. Figure~\ref{fig:loss_comparison} demonstrates this phenomenon across three model families, where fine-tuning consistently reduces loss on benign data while simultaneously decreasing the loss gap on harmful content. Unlike traditional unlearning methods that rely on gradient ascent or parameter noise, our approach leverages second-order information to navigate the parameter space more precisely, enabling targeted modification of safety-relevant parameters while preserving task-specific knowledge.

A key advantage of our method is its ability to exploit the structural separation between harmful and benign regions in the loss landscape, a property that is often absent in standard unlearning scenarios where knowledge is more entangled. This distinct separation allows for more selective restoration than would otherwise be possible. It is worth noting that this favorable landscape structure may not exist in all unlearning contexts, particularly when target and retain knowledge are deeply intertwined. Developing effective curvature-aware unlearning methods for scenarios with less distinct loss landscapes remains an open challenge for future research.

\subsection{Computation Cost}
We evaluate the runtime efficiency of our alignment restoration pipeline (per iteration) by reporting the wall-clock time required for L-BFGS curvature construction and influence-based update steps. All experiments are conducted on a single NVIDIA H100 80GB GPU using 50 harmful samples for influence gradient estimation.

Table~\ref{tab:runtime} summarizes the runtime across three model architectures: LLaMA-2 7B, LLaMA-3 8B, and Qwen 2.5 7B. Despite relying on second-order curvature estimation, our method remains computationally tractable. For example, constructing the L-BFGS history takes approximately 6–7 minutes, and applying the influence update requires only 16–18 seconds. These results demonstrate that our approach is practical and scalable to modern open-source LLMs, with the majority of overhead concentrated in a one-time curvature estimation step.

\begin{table}[h]
    \centering
    \caption{Runtime (in seconds) for curvature construction and influence updating on a single NVIDIA H100 80GB GPU.}
    \label{tab:runtime}
    \begin{tabular}{lcc}
        \toprule
        \textbf{Model} & \textbf{L-BFGS Construction (s)} & \textbf{Influence Updating (s)} \\
        \midrule
        LLaMA-2 7B & 399 & 16 \\
        LLaMA-3 8B & 433 & 18 \\
        Qwen 2.5 7B & 409 & 17 \\
        \bottomrule
    \end{tabular}
\end{table}

%%% Uncomment this line and comment out the ``thebibliography'' section below to use the external .bib file (using bibtex) .




% 	\bibitem{keshet2016prediction}
% 	Keshet, Renato, Alina Maor, and George Kour.
% 	\newblock Prediction-Based, Prioritized Market-Share Insight Extraction.
% 	\newblock In {\em Advanced Data Mining and Applications (ADMA), 2016 12th International 
%                       Conference of}, pages 81--94,2016.

% \end{thebibliography}

\end{document}